\documentclass{article}

\usepackage{arxiv}

\usepackage[utf8]{inputenc} 
\usepackage[T1]{fontenc}    
\usepackage{hyperref}       
\usepackage{url}            
\usepackage{booktabs}       
\usepackage{amsfonts}       
\usepackage{nicefrac}       
\usepackage{microtype}      
\usepackage{lipsum}
\usepackage{graphicx}
\usepackage{amsmath}
\usepackage[ruled,vlined]{algorithm2e}
\usepackage{graphicx}
\usepackage{subfig}
\usepackage{mathtools}
\usepackage{footmisc}
\graphicspath{ {./images/} }

\title{A Fast Anti-Jamming Cognitive Radar Deployment Algorithm Based on Reinforcement Learning}

\author{
 Wencheng Cai \\
  School of Mathematical Sciences\\
  University of Chinese Academy of Sciences\\
  Beijing, China \\
  \texttt{caiwencheng23@mails.ucas.ac.cn} \\
   \And
 Xuchao Gao\\
  Information Science Academy \\
  China Electronics Technology Group Corporation\\
  Beijing, China \\
  \texttt{nossplz@163.com} \\
  \And
 Congying Han\textsuperscript{*}\\
  School of Mathematical Sciences\\
  University of Chinese Academy of Sciences\\
  Beijing, China \\
  \texttt{hancy@ucas.ac.cn} \\
  \And
 Mingqiang Li\textsuperscript{*}\\
  Information Science Academy \\
  China Electronics Technology Group Corporation\\
  Beijing, China \\
  \texttt{limingqiang14@mails.ucas.ac.cn} \\
  \And
 Tiande Guo \\
  School of Mathematical Sciences\\
  University of Chinese Academy of Sciences\\
  Beijing, China \\
  \texttt{tdguo@ucas.ac.cn} \\
}

\begin{document}
\maketitle
\renewcommand{\thefootnote}{}  
\footnotetext{* Corresponding Authors}
\renewcommand{\thefootnote}{\arabic{footnote}}  
\begin{abstract}
The fast deployment of cognitive radar to counter jamming remains a critical challenge in modern warfare, where more efficient deployment leads to quicker detection of targets. Existing methods are primarily based on evolutionary algorithms, which are time-consuming and prone to falling into local optima. We tackle these drawbacks via the efficient inference of neural networks and propose a brand new framework: \textbf{F}ast \textbf{A}nti-Jamming \textbf{R}adar \textbf{D}eployment \textbf{A}lgorithm (FARDA). We first model the radar deployment problem as an end-to-end task and design deep reinforcement learning algorithms to solve it, where we develop integrated neural modules to perceive heatmap information and a brand new reward format. Empirical results demonstrate that our method achieves coverage comparable to evolutionary algorithms while deploying radars approximately 7,000 times faster. Further ablation experiments confirm the necessity of each component of FARDA.
\end{abstract}


\section{Introduction}
Cognitive radar systems have greatly enhanced environment adaptability and target recognition accuracy of radar systems, and have gradually gained attention in recent years. In modern warfare, where electronic countermeasures become increasingly intense, cognitive radars can crack the jamming strategies of enemies and control the electromagnetic network on the battlefield. Besides, in civilian sectors, such as autonomous driving and meteorological monitoring, the adaptive characteristics of cognitive radar can also significantly improve the ability to classify and track targets in complex scenarios. Therefore,  developing cognitive radar technologies can considerably improve radar systems' anti-jamming capacities and information acquisition efficiency, manifesting the cutting-edge of radar development.

Currently, researchers mainly focus on evolutionary algorithms, for instance, Genetic Algorithms (GA) \cite{sun2017fast, wang2019research}, Particle Swarm Optimization (PSO) \cite{yang2015fast, wang2018antenna}, and Artificial Fish Swarm Algorithm \cite{li2024formation} to deploy cognitive radars in the correct positions. However, since evolutionary algorithms require multiple rounds of iterative searches and independent fitness evaluation of the entire population in every round of iteration, they can not solve problems effectively. Furthermore, due to the limitations of the searching strategy, evolutionary algorithms often fail to fully explore the entire solution space, resulting in them easily falling into local optima. These disadvantages directly hinder the process of evolutionary algorithms in finding truly optimal solutions effectively.

Due to efficient inference and extraordinary performance, Deep Reinforcement Learning (DRL) \cite{sutton1998reinforcement} has recently gained attention. Agents trained by DRL showed superior skills in games like Atari \cite{mnih2013playing}, Go \cite{silver2017mastering}, and Starcraft II \cite{vinyals2019grandmaster}. DRL is also used to tackle real-world scenarios, including autonomous driving \cite{kiran2021deep}, math problems \cite{shao2024deepseekmath}, and machine translation \cite{feng2025mt}. Also, researchers use RL to solve combinatorial optimization problems \cite{bengio2021machine}, such as Traveling Salesman Problem \cite{kool2018attention} and Mixed Integer Linear Programming Problem \cite{tang2020reinforcement}.

Recently, combining DRL methods with radar problems has attracted increasing attention from researchers. Zhu et al. \cite{zhu2025resource} used a hybrid action space RL to enhance the resource utilization of the Multiple-Input Multiple-Output (MIMO) radar system. Jiang et al. \cite{jiang2017optimal} formed a Markov Decision Process (MDP) and used a low complexity joint optimization algorithm to manage radars’ resources. Yang et al. \cite{yang2023multi} used Q-learning to solve the resource scheduling problem of radars. Hao et al. \cite{hao2020application} used Q-learning and SARSA algorithm to select the radar’s anti-jamming policy method. Jiang et al. \cite{jiang2023intelligent} and Wang et al. \cite{wang2025} applied Deep Deterministic Policy Gradient algorithm to select the radar’s anti-jamming policy. Zhu et al. \cite{zhu2023antenna} used Proximal Policy Optimization (PPO) algorithm to choose the location of antennas of an MIMO radar system in a jamming-free situation. To the best of our knowledge, no research has been conducted on the anti-jamming radar deployment problem using DRL.

In this paper, we innovatively introduce DRL method into the radar anti-jamming deployment problem: \textbf{F}ast \textbf{A}nti-Jamming \textbf{R}adar \textbf{D}eployment \textbf{A}lgorithm (FARDA). We first model the deployment problem as a combinatorial optimization problem and propose the corresponding mathematical model. However, two factors impede us from solving this problem quickly and accurately. The first factor is the complicated optimization objective, while the second factor is the large search space. Therefore, we analyze the characteristics of the problem, extract inherent properties, and leverage them to optimize the problem's modeling, including environment modification, dimensionality reduction, and constraint relaxation. These transformations reduce the computational complexity of the optimization functions and shrink the search space, making the problem easier to solve. Next, we formulate the Markov Decision Process (MDP) for the corresponding problem and propose a DRL framework, where we specifically design an encoder in our policy network. Not only can it extract the heatmap information of the detection probabilities, but it can also memorize the temporal features from previous states. In addition, we introduce a brand new reward shaping method: \textbf{C}onstraint \textbf{V}iolation \textbf{D}egree \textbf{P}enalty and \textbf{EXP}onential Function \textbf{R}eward (CVDP-EXPR), which can better guide the agent in the learning stage. 

Finally, sufficient numerical experiments are conducted to show that our FARDA achieves coverage competitive to evolutionary algorithms while deploying radars approximately 7,000 times quicker. Also, our algorithm outperforms evolutionary algorithms by a factor of 120 in the efficiency metric we have designed. We also conduct a series of ablation tests to demonstrate the necessity of the encoder module in FARDA and the CVDP-EXPR we proposed. 

Our contribution can be summarized as follows:

\begin{itemize}
    \item We formulate the anti-jamming cognitive radar deployment problem as a combinatorial optimization problem, and point out some obstacles to solving the problem.
    \item We design a new framework, FARDA, to tackle the problem. We first explore the characteristics of the problem and modify it to make it easier to solve. Then, we specifically design a DRL framework to address the modified combinatorial optimization problem. We also devise CVDP-EXPR in FARDA, benefiting its training and deployment.
    \item Numerical experiments show that our method outperforms evolutionary algorithms in both speed, which is approximately 7000 times faster, and coverage area. Also, ablation experiments demonstrate the necessity of each part of our proposed encoder in FARDA and the design of CVDP-EXPR.
\end{itemize}

\section{Problem Formulation}

The main objective for the anti-jamming deployment of cognitive radar is as follows: Several jamming nodes deployed by the enemy will appear at discrete points set $D_J$ within a two-dimensional region $S$. We should deploy several cognitive radars at discrete points set $D_R$ within another area of $S$.

Formally, region $S$ is a $20 $km$\times 120 $km area, to simplify the description, we place it in a rectangular coordinate system:

\begin{equation}
\begin{aligned}
    S &= [3 \times 10^4,\, 5\times 10^4]\times [0,\, 1.2\times 10^5],\\
    D_{J} & = ([4 \times 10^4,\, 5\times 10^4] \times [6\times 10^4,\, 1.2 \times 10^5])\cap \mathbb{Z},\\
    D_{R} &=( [3 \times 10^4,\, 4\times 10^4] \times [0,\, 6\times 10^4])\cap \mathbb{Z}.\\
\end{aligned}
\end{equation}

We should maximize the radars' detection region $\tilde{S}$ in $S$. However, it is difficult for a computer to calculate the region $\tilde{S}$ for two reasons: first, the shape of $\tilde{S}$ may be highly irregular; second, processing continuous data on computers is extremely difficult, so we need to discretize it. To match the precision of preceding tasks, we discretize the surveillance region $S$ by sampling a point every 100 meters to obtain the discrete matrix $D$, where we denote the counts of rows and columns as $n$ and $m$, respectively. Furthermore, we denote $\tilde{D}$ the detected points by radars on $D$.

\begin{figure}[t]
\centering
\includegraphics[width=0.85\textwidth]{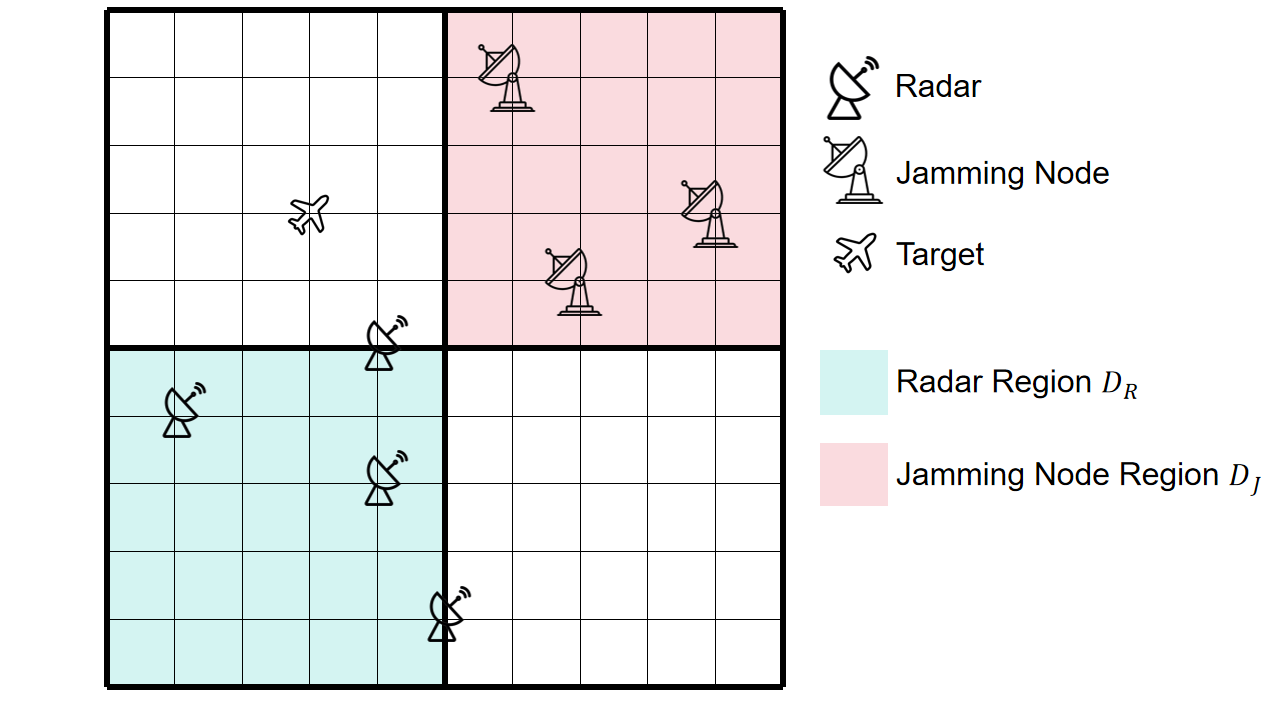}
\caption{An example figure of the problem.}
\label{fig:problem}
\end{figure}

Figure~\ref{fig:problem} shows an example of the problem. The grid in light red represents $D_J$, and the grid in light green represents $D_R$. The three radars in $D_J$ are enemy jamming nodes, while the four radars in $D_R$ are our deployed radars. The aircraft represents a possible location of the target. Note that this figure has been scaled proportionally: the actual area is a rectangle. Furthermore, the grid in the figure is just an example; the actual grid will be denser.

In area $D_J$, suppose the enemy has placed a jamming node combination $J$ and its coordinates are described as follows:

\begin{align}
J = (j_1,\cdots, j_{|J|} ),\; j_i = (j_{i_x}, j_{i_y})^T \in D_{J} , \, i = 1, \cdots , |J|.
\end{align}

Assume that we have deployed a radar combination $R$ in area $D_R$ and denote the coordinates as follows:

\begin{align}
R = (r_1,\cdots, r_{|R|} ),\; r_i = (r_{i_x}, r_{i_y})^T \in D_{R} , \, i = 1, \cdots , |R|.
\end{align}

We use four steps to calculate $\tilde{D}$. Note that we have omitted the superscripts and subscripts from some of the formulas where there is no ambiguity to simplify the formulas.

\begin{itemize}
    \item Step 1: calculate the detection probability of a single radar countering a single jamming node at a certain point.

    For a radar $r = (r_x, r_y)^T \in R$, a jamming node $j = (j_x,j_y)^T \in J$ and a certain point in discrete surveillance area $ d = (d_x, d_y)^T \in D$. We can calculate relative coordinates $\delta_j =(\delta_{j_x}, \delta_{j_y})^T$, distance $R_j$ and angle $\theta_j$ of $r$ relative to $j$:
    
    \begin{equation}
    \begin{aligned}
        \delta_{j} &= (r_x - j_{x}, r_y - j_{y})^T, \\
        R_{j} &= \Vert \delta_{j} \Vert_2,\\
        \theta_{j} &= \begin{cases}
            \arccos({\delta_{j_x}}/{R_{j}}) & \text{if }\, r_y > j_{y}, \\
            2\pi - \arccos({\delta_{j_x}}/{R_{j}}) & \text{else}.
        \end{cases}
    \end{aligned}
    \end{equation}
    
    Similarly, we can calculate relative coordinates $\delta_d = (\delta_{d_x}, \delta_{d_y})$, distance $R_d$ and angle $\theta_d$ of $r$ relative to $d$:

    \begin{equation}
    \begin{aligned}
        \delta_{d} &= (r_x - d_{x}, r_y - d_{y})^T, \\
        R_{d} &= \Vert \delta_{d} \Vert_2,\\
        \theta_{d} &= \begin{cases}
            \arccos({\delta_{d_x}}/{R_{d}}) & \text{if }\, r_y > d_{y}, \\
            2\pi - \arccos({\delta_{d_x}}/{R_{d}}) & \text{else}.
        \end{cases}
    \end{aligned}
    \end{equation}

    Next, we calculate the effective power that can be received by radar $r$ after the echoes of the signals emitted by the radar $r$ and jamming node $j$ at point $d$, which we denote as $P_r$ and $P_j$ respectively:

    \begin{equation}
    \label{eq:power}
    \begin{aligned}
        P_{r} &= \frac{\tilde{P}_rG_TG_R\lambda^2K^2B^2}{(4\pi)^3(R_d)^4},\\
        P_{j} &= \frac{\tilde{P}_j\lambda^2KB}{(4\pi)^2 (R_{j})^2 F_r}.
    \end{aligned}
    \end{equation}

    In \eqref{eq:power}, $\tilde{P}_r$ and $\tilde{P}_j$ are the power of transmit antenna of radar $r$ and jamming $j$, $G_T$ and $G_R$ are the gain of the transmit and receive antennas of radar $r$, $\lambda$ is the wavelength of the radar electromagnetic wave, $K$ is the array element count in the radar antenna, $B$ is the bandwidth of radar, and $F_r$ is the pulse repetition rate.

    Next, we give an angle threshold $\theta$, and calculate whether there is a jamming node within the sector region centered at $r$ with central angle $[\theta_d - \theta, \theta_d + \theta]$, and thereby calculate the Signal to Interference Plus Noise Ratio (SINR):
    \begin{align}\label{eq:SINR} SINR = \begin{cases}
    {P_r}/ P_{j} & \text{if}\  \theta_{j}\in [\theta_d - \theta,\, \theta_d + \theta], \\
    {P_r}/{P_n} &  \text{else}.\\
    \end{cases}
    \end{align}

    $P_n$ denotes the power of noise in \eqref{eq:SINR} and can be determined as follows:
    \begin{align}
        P_{n} = \frac{K_eT_0KB^2F_e}{F_r},
    \end{align}

    where $K_e$ is the Boltzmann constant, $T_0$ is the room temperature, and $F_e$ is the noise factor.

    Finally, we set the false alarm rate $Pr_{fa} \ll 1$, and the detection probability of radar $r$ against jamming node $j$ at point $d$ is:
    \begin{align}
    \label{eq:prob}
    Pr = (Pr_{fa})^{1 / (1+SINR)}.
    \end{align}

    \item  Step 2: Calculate the detection probability of a single radar against a jamming combination at a certain point. 

    Most of the calculating processes for the jamming node set $J$ are similar to step 1, except for calculating the SINR in \eqref{eq:SINR}. Specifically, for each jamming node $j_i\in J$, we can calculate its angle $\theta_{j_i}$ with the radar $r$ and its power $P_{j_i}$ at point $d$. We then calculate the interference set $C$ that falls within the sector region:
    \begin{equation} C = \{i \,|\, \theta_{j_i}\in [\theta_d - \theta,\, \theta_d + \theta] , \, i = 1,\, \cdots, \,|J|\}.\end{equation}

    Then SINR of radar against jamming combination at point $d$ can be calculated using the following formula:
    \begin{align} SINR = \begin{cases}
    {P_r}/{\sum_{i \in C} P_{j_i}} & \text{if } \left\lvert C \right\rvert \ne 0, \\
    {P_r}/{P_n} &  \text{else}.\\
    \end{cases}
    \end{align} 

    Then, using \eqref{eq:prob}, we can calculate the detection probability of a radar $r$ countering the jamming node combination $J$ at point $d$, which we denote as $Pr^J$.

    \item Step 3: Calculate the detection probability of a radar combination against a jamming node combination at a certain point. 

    For a radar combination $R$, we can calculate every single radar $r_i \in R$ as the procedure of step 2, and obtain a detection probability of radar $r_i$ against the jamming combination $J$, denoted as $Pr_i^J$. Then, the detection of $R$ against $J$ can be calculated as:
    \begin{align}
        Pr^{R} = 1-\prod_{i = 1}^{|R|} (1-Pr^J_i).
    \end{align}

    \item Step 4: calculate the detected set $\tilde{D}$ of radar combination against jamming node combination in set $D$.

    In the surveillance area set $D$, each point $d$ corresponds to a detection probability $Pr^R_d$ calculated by step 3. Given a threshold $\tau$, when the detection probability $Pr_d^R \ge \tau$, we consider that if a target appears at point $d$, the radar combination can detect that point. Formally, the set of points detected by the radar combination under jamming nodes can be characterized as follows:
    \begin{align}
    \label{eq:A area}
        \tilde{D} = \{d \in D | Pr^{R}_d \ge \tau\} .
    \end{align}
    
\end{itemize}

We aim to cover as much of the surveillance area $\tilde{D}$ as possible. Therefore, we propose the following combinatorial optimization problem:
\begin{align}
\label{eq:objective}
\begin{split}
    \max & \,\,\,\,\, \vert\tilde{D}\vert / \vert D\vert\\
    s.t. & \,\,\,\,\,  r_i \in D_R, \, i = 1, \cdots, |R|
\end{split}
\end{align}

In \eqref{eq:objective}, $\tilde{D}$ is the detected discrete set, $D$ is the discrete region set sampled in 100 meters, $r_i$ is the radar position, and $D_R$ is the radar deployable set.

As can be seen from the above modeling process, calculating the objective function in this combinatorial optimization problem is extremely complex. We summarize the detailed difficulties of this problem as follows: 

\begin{itemize}
    \item The optimization functions have some undesirable properties: First, compared to jamming-free conditions, our scenario contains the variables of jamming positions, resulting in a higher parameter dimension in the optimization function; Second, the function is highly sensitive and coupled in radar and jamming factors, a slight change in positions can cause significant fluctuation of the final value in the function.

    \item Discontinuity property of the optimization function: The existence of the jamming nodes causes the \eqref{eq:SINR} to change from continuous to discontinuous. Discontinuous functions have many undesirable properties compared to continuous functions, making the problem more challenging.

    \item Large decision space: Since all possible combinations of radars can be represented by the combination number $C(|D_R|, |R|)$, when $|R|$ is much smaller than $|D_R|$, this combination number can be approximated by an exponential function. This indicates that the size of our decision space is exponential, with its base $|D_R|$ reaching the order of $10^8$. This means that the decision space for this problem is enormous, making it impossible to solve using traditional search algorithms. Additionally, using evolutionary algorithms will likely lead to local optima, causing the algorithm to converge to a suboptimal solution. 
\end{itemize}

\section{Methodology: FARDA}

\begin{figure*}[!t]
\centerline{\includegraphics[width=\textwidth]{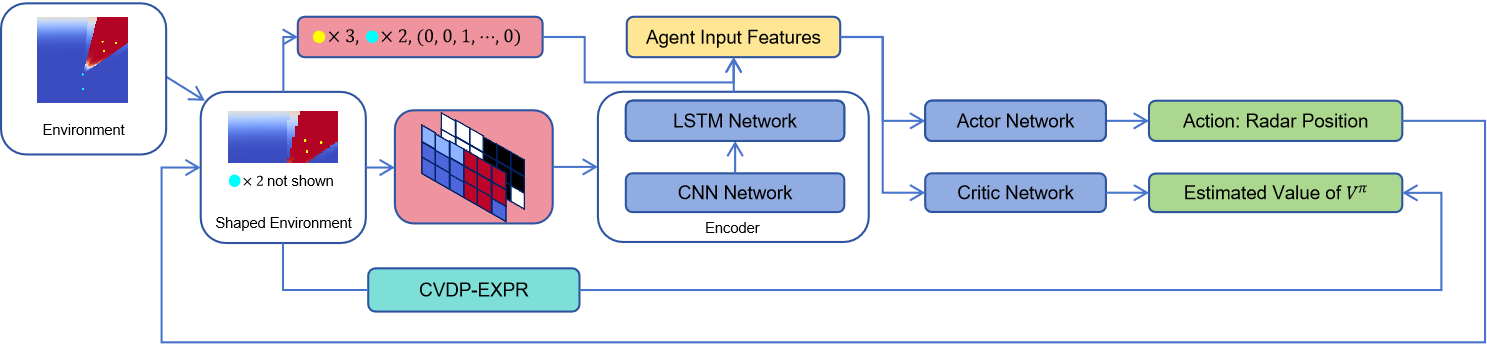}}
\caption{The formulation FARDA, concluding environment shaping, encoder, actor, and critic network, and CVDP-EXPR.\label{fig:example}}
\end{figure*}

We have designed a new method, FARDA, to address this problem, which is summarized in Figure~\ref{fig:example}. We will introduce FARDA step by step in this section, starting with problem simplification, MDP modeling, policy network, and finally, our reward shaping method: CVDP-EXPR.

\unskip

\subsection{Problem Simplification}

To make this highly complex combinatorial optimization problem easier to solve, we simplify it using a series of approaches, including dimension reduction, relaxation, and environment simplification.

\subsubsection{Dimension Reduction}

For the original problem, we first tried PSO and GA on it, and found that radars on the boundary of $D_R$ contribute most of the coverage. So we conceive the idea of trying to place radar only on the boundary, and conduct an experiment to see the coverage difference of placing radars only on the boundary and placing them in the region.

We first define the boundary $B_R$ of the deploy region as follows:
\begin{equation}
\begin{aligned}
    &B_{up} = (t_{up}, 6\times 10^4),  &3\times 10^4 \le t_{up} \le 4\times 10^4, \\
    &B_{right} = (4\times 10^4, t_{right}), &0\le t_{right} \le 6\times 10^4,
\end{aligned}
\end{equation}
\begin{equation}
    \begin{split}
        \label{eq:br}
    B_R = (B_{up} \cup B_{right}) \cap D_R,
    \end{split}
\end{equation}
where, intuitively, $B_{up}$ and $B_{right}$ represent the upper and right boundaries of $D_R$, while $B_R$ is the set of all points on the upper and right boundaries of the region $D_R$.

For the original optimization problem \eqref{eq:objective}, we attempted to deploy radars using GA and PSO in regions $D_R$ and $B_R$. Descriptions of the GA and PSO are provided in Appendix \ref{sec:GA} and \ref{sec:PSO}, while the hyperparameters are detailed in Section \ref{sec:experiments setups}. To simplify our descriptions, we denote the particle swarm and genetic algorithms deployed on $B_R$ as PSO1D and GA1D, respectively, and denote the algorithms deployed on $D_R$ as PSO and GA. We randomly select 100 sets of jamming node combination locations, run these algorithms, and record the average coverage of radar combinations of different algorithms on these 100 sets. The results are shown in Table \ref{tab:1d vs 2d}.

The experiment results indicate that when using evolutionary algorithms, the average performance along the boundary $B_R$ consistently outperforms performance in the deploy area $D_R$. A possible factor is that although deploying along $B_R$ may discard some more optimal solutions than deploying in $D_R$, the corresponding reduction in search space facilitates algorithm convergence. Moreover, the performance loss from the discarded solutions is outweighed by the efficiency gains in convergence achieved through the reduced search space. Therefore, we will also deploy the radar along $B_R$ in our subsequent reinforcement learning algorithms.

\begin{figure}[!t]
\centering
\subfloat[\centering PSO Algorithm]{\includegraphics[width=0.45\textwidth]{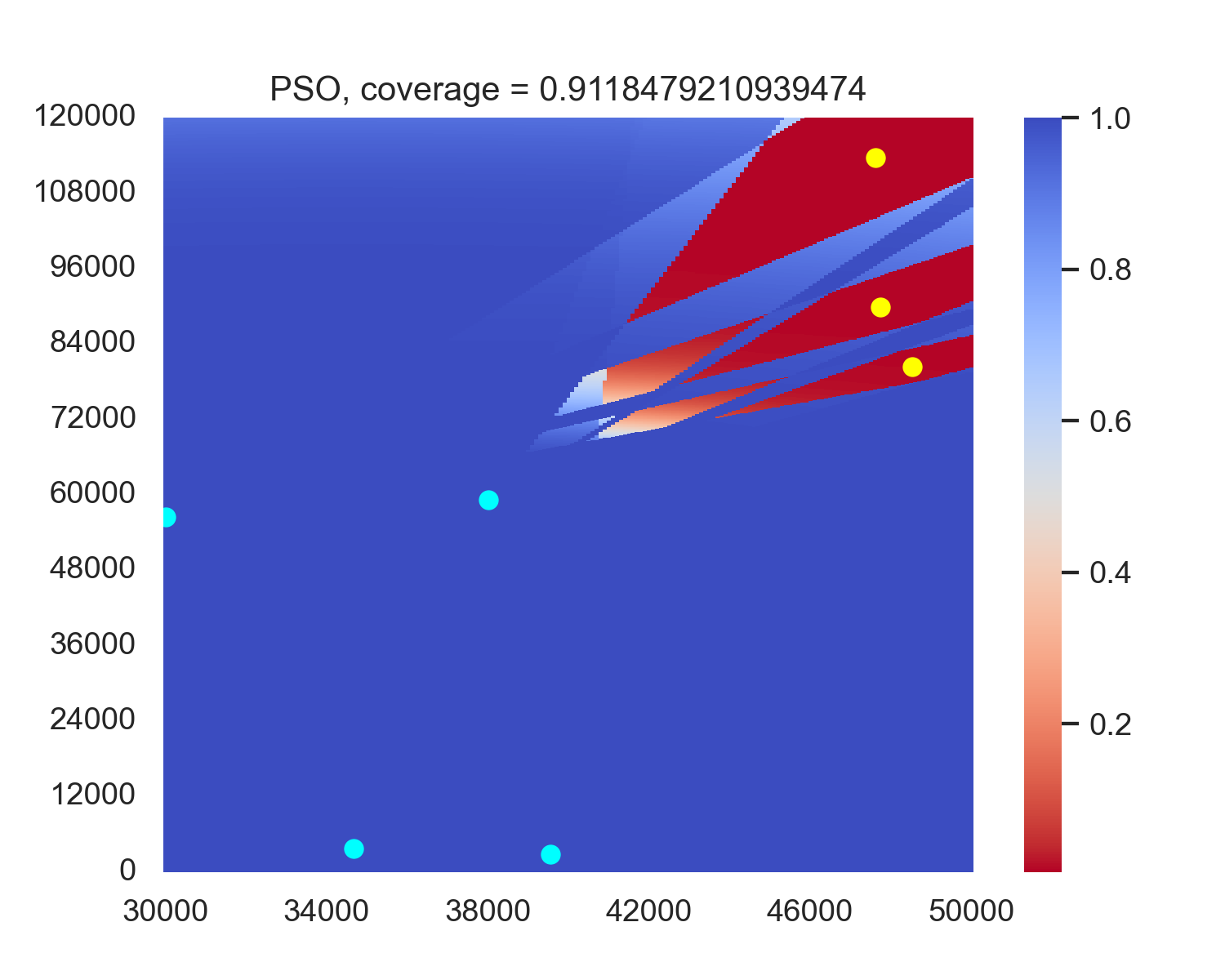}}
\subfloat[\centering GA algorithm]{\includegraphics[width=0.45\textwidth]{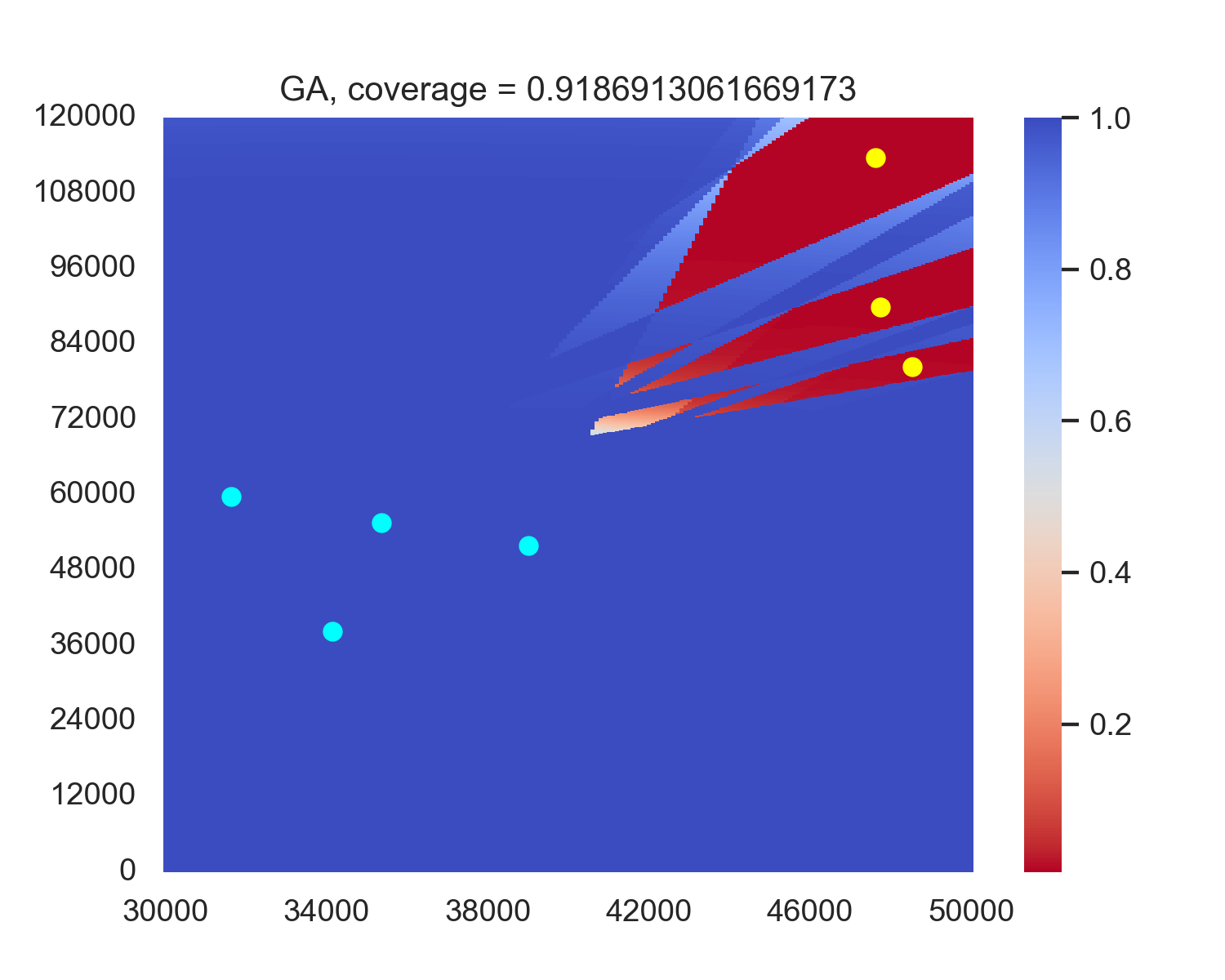}}\\
\subfloat[\centering PSO1D Algorithm]{\includegraphics[width=0.45\textwidth]{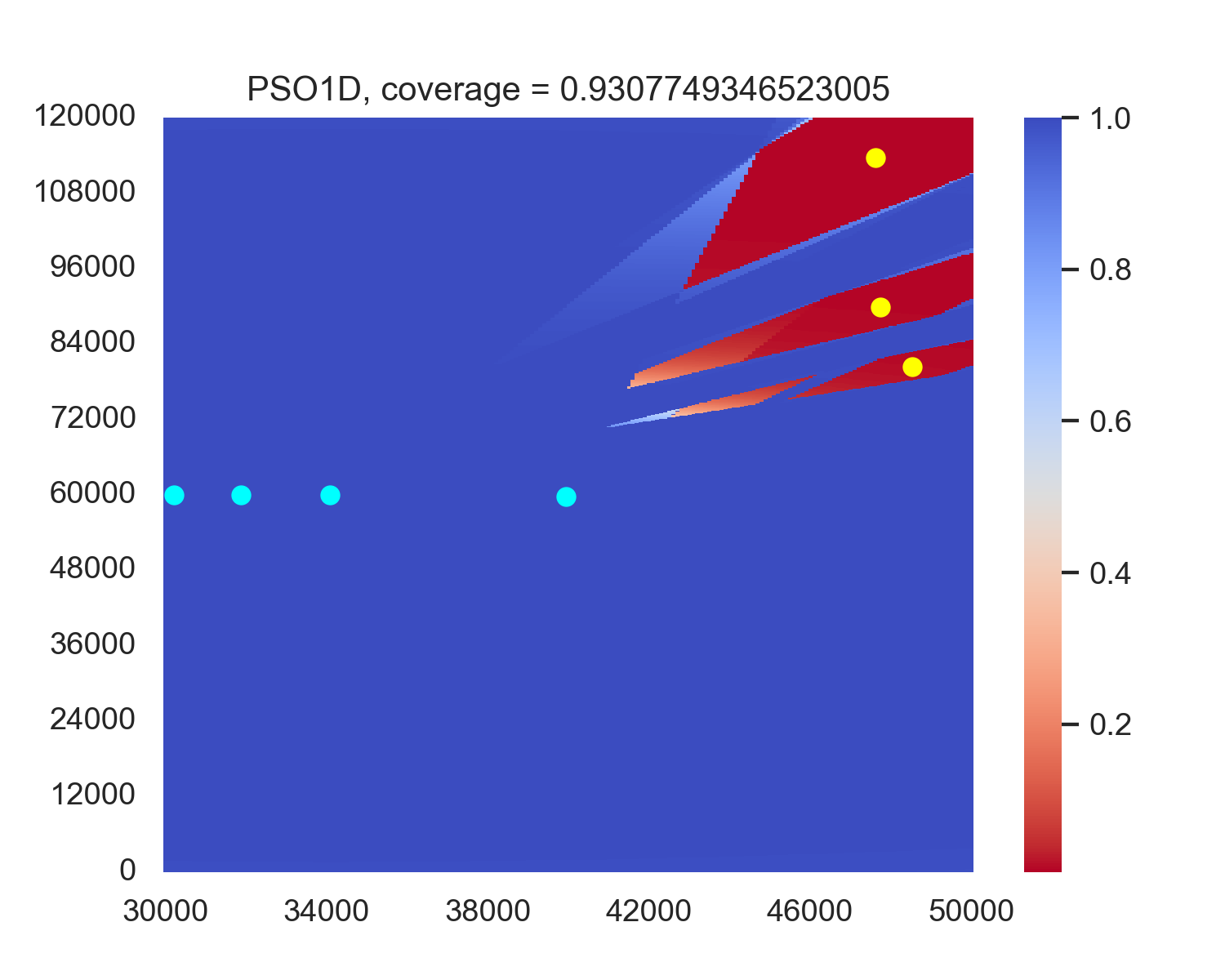}}
\subfloat[\centering GA1D Algorithm]{\includegraphics[width=0.45\textwidth]{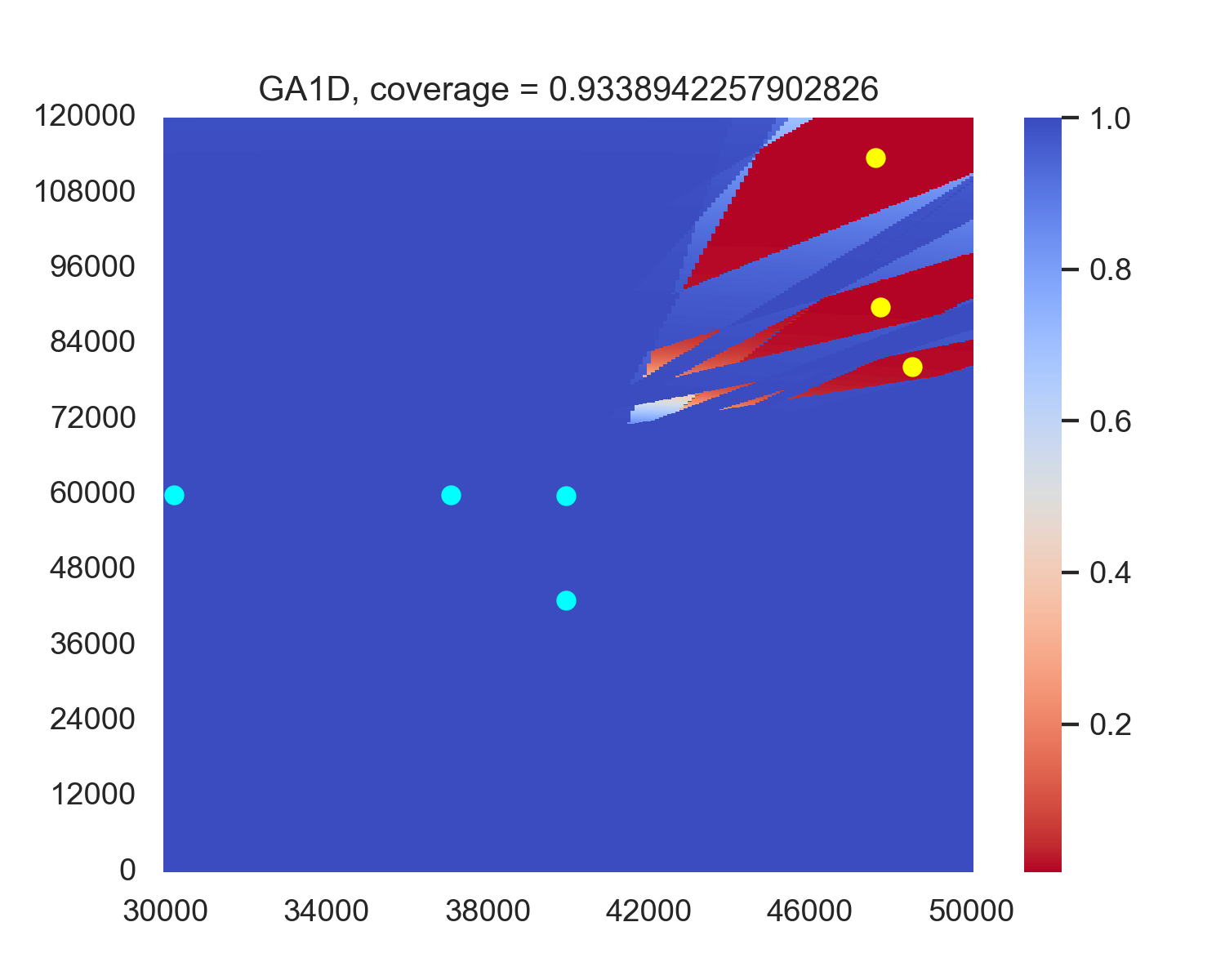}}
\caption{When using PSO and GA, the results obtained from deploying within the deploy area $D_R$ versus along its boundary $B_R$. Correspondingly: (a) using PSO and deploy in $D_R$, (b) using GA and deploy in $D_R$, (c) using PSO and deploy in $B_R$, (d) using GA and deploy in $B_R$. The four cyan dots indicate radar positions, while the upper three yellow dots represent jamming node positions. The color of the heatmap ranges from blue, indicating a detection probability greater than 0.5, to red, indicating a detection probability less than 0.5.\label{fig:1D vs 2D}}
\end{figure}

Figure \ref{fig:1D vs 2D} shows an example case of the performance in PSO and GA versus PSO1D and GA1D, showing that PSO1D and GA1D outperform PSO and GA by around 0.015 of coverage in this example. An intuitive sense is in PSO and GA, only the radars near the boundary contribute most of the coverage, while every radar in PSO1D and GA1D contributes to the final coverage. A possible reason for the non-boundary radars of PSO and GA is that this position combination becomes a local optimum, and non-boundary radars are stuck in this position and cannot get out.

\begin{table}
\centering
\caption{The performance gap between deploying radars using evolutionary algorithms along $B_R$ and within $D_R$. The \underline{underlined} part indicates the best result.}
\begin{tabular}{c|c|c}
    \hline   Methods & Average Coverage (\%) & Gap \\   
    \hline   
            PSO $(D_R)$& 92.70 & 0.79\% \\
            GA $(D_R)$ & 92.30 & 1.22\% \\
    \hline
            PSO1D $(B_R)$& 93.33 & 0.12\% \\
            GA1D $(B_R)$& \underline{93.44} & - \\
    \hline   
\end{tabular}
\label{tab:1d vs 2d}
\end{table}

\subsubsection{Relaxation}

In this problem, we try to set the action space on discrete point set $B_R$ and find it hard for the agent in learning and converge. A further analysis shows that this action space may be enormous in our training environment. Therefore, we relax the problem by transforming the agent's action space into a continuous region boundary.

Specifically, we define the following continuous boundary $B_R'$ and set the agent's action space as the coordinates of $B_R'$. After the agent selects a radar deployment location, we move the position nearest to $B_R$ as the final radar location for agent-environment interaction. This transformation converts the agent's action space from discrete to continuous. Agents will make decisions on coordinates instead of a set of specific points, significantly reducing the dimensionality of the action space, making it easier to deploy. 
\begin{equation}
    \begin{split}
    B_R' = B_{up} \cup B_{right}.
    \end{split}
\end{equation}

\subsubsection{Environment Simplification}

During the implementation of the optimization function, we found that the calculation speed of this function is extremely slow, taking about 15 seconds. Our further analysis reveals some possible reasons. The first is that in the optimization function, we need to calculate the detection probability on approximately $200\times 1200 = 2.4\times 10^5$ points, which is numerous and redundant for the problem. Second, the detection function contains a lot of calculations of trigonometric functions, which may further decrease the speed of calculation.

To address this trouble, we define a new area $D'$ for training the agent: in this area, we remove the lower half of $D$ and change the sampling interval to 500 meters. We remove the lower half of the heatmap because radars can consistently detect these parts, so this region carries much less information than the upper half. We then use this $D'$ to calculate the detected point set $\tilde{D}'$ by the radar combination, and obtain a detected area $|\tilde{D}'|/|D'|$ for the reward given by the environment. Note that this does not alter the optimization objective mentioned in the optimization problem \eqref{eq:objective}; we merely employ a new environment for training.

\begin{figure}[!t]
\centering
\subfloat[\centering Sampling in 100 meters]{\includegraphics[width=0.45\textwidth]{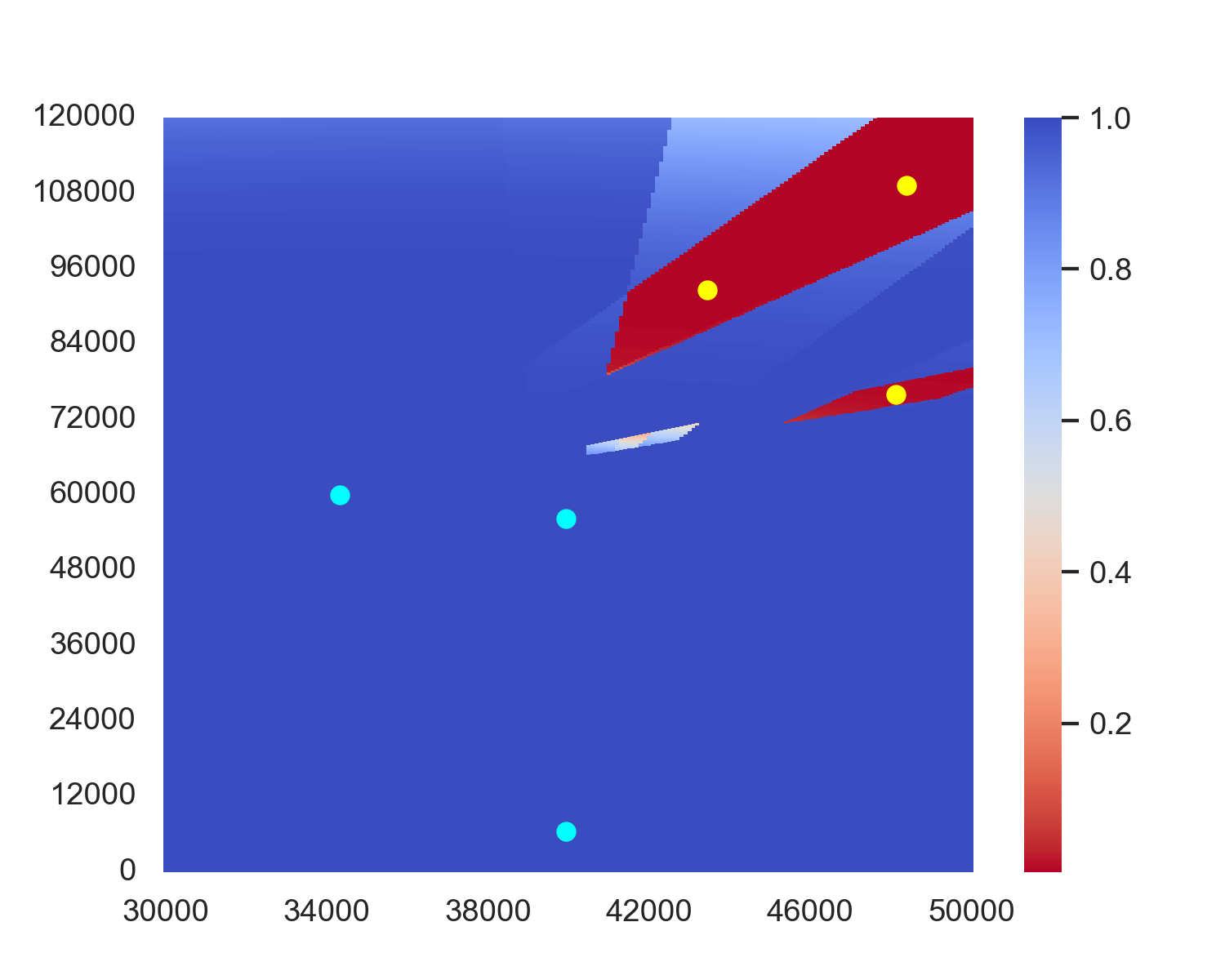}}
\subfloat[\centering Sampling in 500 meters]{\includegraphics[width=0.45\textwidth]{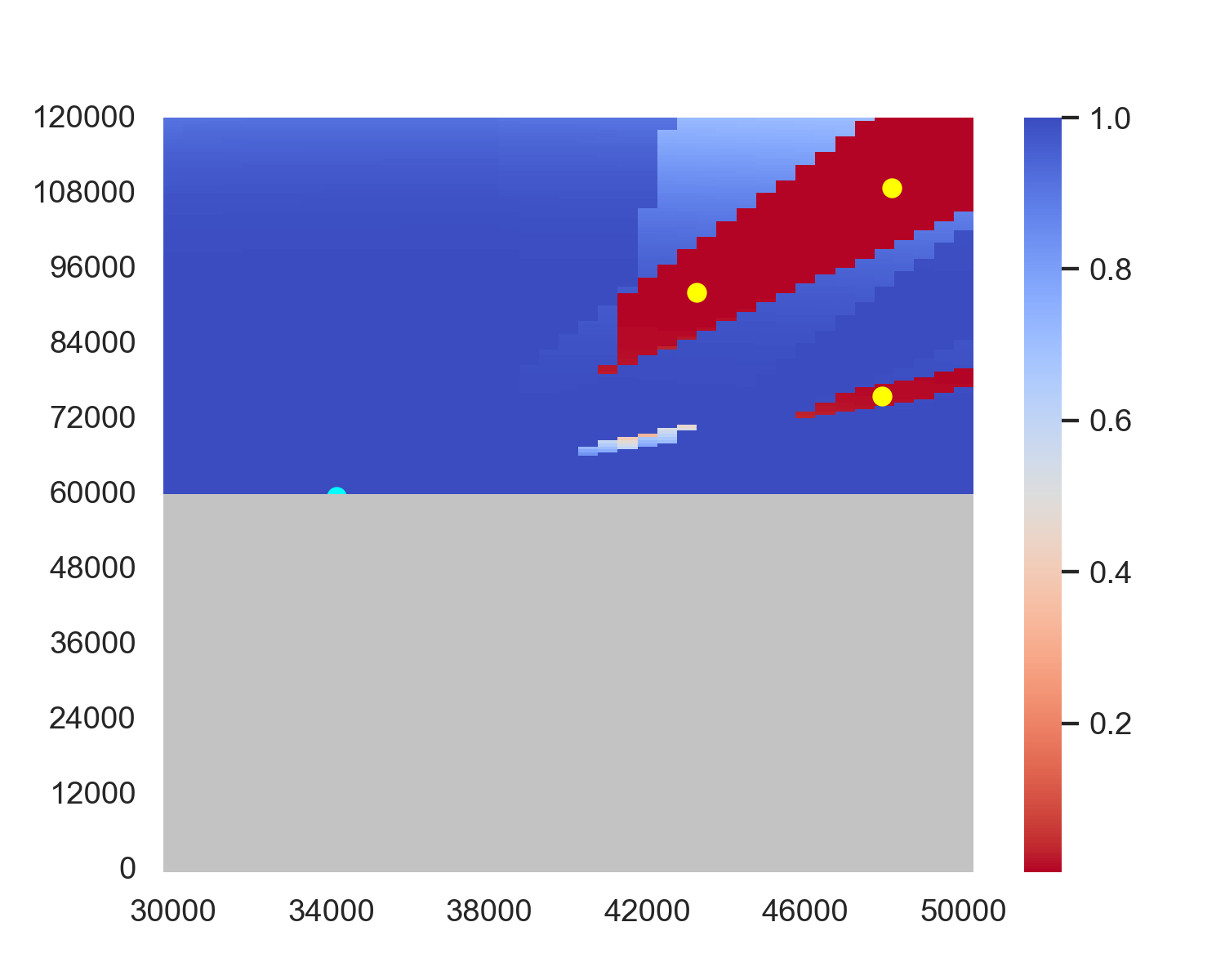}}
\caption{(a) The heatmap of sampling points in 100 meters. (b) The heatmap of sampling points in 500 meters. The gray rectangle region of (b) denotes the region we will discard. Note that in (b), the positions of the two radars are not shown.\label{fig:environment simplification}}
\end{figure}

Figure \ref{fig:environment simplification} shows a three-radar example of the heatmap before and after environment simplification. From the figure, it is evident that the shape of the undetected area remains similar. Besides, the detection probability in the lower half is essentially 100\% and does not carry much information. Therefore, our operation is reasonable.

Suppose the size of $D'$ is $n' \times m'$, the relationship of shape between $D$ and $D'$ will be: $n' = 1/10n,\, m' = 1/5 m$. So we reduce 50 times fewer points when calculating the coverage area by sacrificing a small amount of heatmap information, significantly decreasing computation time and accelerating training speed.

\subsection{Formulation of MDP}

We transformed the radar deployment problem into a sequential decision-making problem. Specifically, at time step $t$, we determine the deployment location of the $t$th radar based on the positions of already deployed radars, jamming node locations, and environmental information. We further formulate it as an MDP $\mathcal{M} = (\mathcal{S}, \mathcal{A}, \mathcal{T}, \mathcal{R},\gamma,  \mu_0)$:

\begin{itemize}
    \item State space $\mathcal{S}$: At time step $t$, the state comprises: the current detection probability heatmap of $D'$, the location of jamming nodes, historical actions (i.e., positions of radars already deployed), and a one-hot vector indicating which radar should be deployed next. That is:

    \begin{align}
    \label{eq:state}
        s_t = ( M_t, \, J, \, R_t ,\, v_t ).
    \end{align}

    Formally, $M_t \coloneqq \{Pr^{R}_{d, t}\}_{d\in D'}$ represents the heatmap of detection probabilities in $D'$ given the combination of radar positions output by the agent before time $t$. $J$ represents the jamming node combination. $R_t$ is a vector with $2\times |R|$ dimensions, where the first $2\times (t-1)$ dimensions are the radar's historical positions, and the remaining dimensions are set to zero. $v_t$ denotes the one-hot vector with dimension $|R|$, with the $t$th dimension is set to one.

    \item Action space $\mathcal{A}$: At time step $t$, the agent outputs an action $a_t$ as a two-dimensional coordinate $a_t = (a_{x,t}, a_{y,t})$, indicating the position of the $t$th radar placement.

    \item State transition $\mathcal{T}$: After the agent takes action $a_t$ at time step $t$, a deterministic transition occurs. First, the environment computes the new detection probabilities for each point, generating a new probability heatmap $M_{t+1}$, along with a new action history $R_{t+1}$ and a new one-hot vector $v_{t+1}$. These form a new state $s_{t+1} = \{ M_{t+1}, \, J, \, R_{t+1} ,\, v_{t+1} \}$

    \item Reward function $\mathcal{R}$: Our goal is to maximize the detected area as shown in equation \eqref{eq:objective}. After each interaction with the environment, the environment provides a reward equal to the coverage ratio $|\tilde{D}'|/ |D'|$ calculated from all deployed radars.

    \item Discount factor $\gamma$: We set $\gamma = 1$ in this MDP due to the limited horizon length in the environment of this problem.

    \item Initial distribution $\mu_0$: Before each episode starts, we randomly sample jamming nodes' location $J$ from the environment. Since no radar is initially deployed, we initialize the probability matrix $M_0$ as a zero matrix. Then, we initialize $H_0$ as a zero vector and $v_0$ as a one-hot vector where the first dimension is set to 1. Based on this, we can obtain the initial state $s_0 = (M_0,J,R_0,v_0)$.

\end{itemize}

\subsection{Policy Network}

The policy network in FARDA is based on an encoder network, along with an actor and a critic network. We will describe these networks in detail in this section.

\subsubsection{Encoder Network}

Due to the presence of the detection probability heatmap, it is necessary to extract its features. Furthermore, as we need to deploy radars sequentially, the network needs to remember the newly gained area between two consecutive radar deployments. 

To deal with these two problems, we have designed our encoder network, incorporating a Deep Convolutional Neural Network (DCNN) and a Long-Short Term Memory (LSTM) block. The primary purpose of the encoder is to abstract the heatmap information into a feature vector using DCNN, then employing LSTM to remember the area changes by the newest deployed radar. We will introduce these two networks separately:

\begin{itemize}
    \item The DCNN block: 
    
    In the state set \eqref{eq:state}, the DCNN block only uses the heatmap information $M_t$ as input. The primary objective of this DCNN block is to extract heatmap information into a latent space and learn a favorable embedding.

    First, we add a new channel to the heatmap. Formally, for every $Pr_{d, t}^{R} \in M_t$, we perform the following operation and add the derived $M_t' \coloneqq\{Pr_{d, t}'^{R} \}_{d\in D'}$ as a new channel.
    \begin{align}
    \label{eq:one more channel}
    Pr'^{R}_{d, t} = \begin{cases}
        0 & \text{if} \; Pr^{R}_{d, t} < \tau \\
        1 & \text{if} \;Pr^{R}_{d, t} \ge \tau.
    \end{cases}
    \end{align}

    This new channel aims to amplify the influence of $\tau$ while enabling the network to learn the threshold $\tau$ we have set.

    Next, we put  $M_t$ and $M_t'$ as two channels, denoted as $X_{t,0}^C$, into the CNN, which can be concretely expressed as follows:
    \begin{align}
    X_{t, 0}^{C} = [M_t, M_t'] \in \mathbb{R}^{2\times n'\times m'}.
    \end{align}
    
    For the convolutional layer and pooling layer, we can use the following formula to represent the forward propagation at time step $t$:
    \begin{align}
    X_{t,j}^{C} = \text{Maxpool}(\sigma(X_{t,{j-1}}^{C} * K_j^{C})), \, j = 1, 2, 3,
    \end{align}
    
    where $K_j^{C}$ denotes convolutional kernel, $\sigma$ denotes the sigmoid activate function. The parameters of convolutional kernels and pooling layers are given in Table \ref{tab:CNN kernels}.

    \begin{table}
    \centering
    \caption{The parameters of convolutional kernels and pooling layers.}
    \begin{tabular}{p{25pt}|p{95pt}|p{95pt}}
        \hline   $j$ & Kernel Size & Pooling Size \\   
        \hline   1 & $ 5\times 5\times 2\times 6$ & $2\times 2$ \\
                 2 & $ 3\times 3\times 6\times 16$ & $2\times 2$ \\ 
                 3 & $ 3\times 3\times 16\times 10$ & $2\times 2$ \\
        \hline   
    \end{tabular}
    
    \label{tab:CNN kernels} 
    \end{table}

    After the convolutional and pooling operations, we can obtain a two-dimensional embedding $X^C_{t,3}$. We flatten the embedding to one dimension, denoted as $X^C_{t,4}$. Then we use a multilayer perception (MLP) to obtain the final embedding vector $H^C_t$ by DCNN, where the hidden sizes of the three fully connected layers are 128, 64, and 64, respectively, and the activation function is the sigmoid function. We can use the following formula to present this process more formally:
    \begin{align}
    H^{C}_t = MLP_c(X_{t,4}^{C}) \in \mathbb{R}^{64}.
    \end{align}
    
    \item The LSTM block:

    Before every episode starts, the LSTM block initializes two zero vectors $h_0$ and $c_0$ as the hidden vectors, which can be formalized as follows:
    \begin{align}
    \label{eq:lstminit}
        h_0=c_0= \textbf{0}^{64}.
    \end{align}
    
    At time step $t$, when the embedding vector of the DCNN block $H^{C}_t$ passes through the LSTM block, the hidden vectors obtained from the last time step $h_{t-1}$ and $c_{t-1}$ will become the input hidden vectors for the LSTM block. After these three vectors pass through the LSTM block, three more vectors $X^{L}_t$, $h_t$, $c_t$ will be obtained. We will then feed $X^{L}_t$ to an MLP with one hidden layer of size 64 to get the output vector $H^L_t$. The formula below shows the detailed process of the LSTM block:
    \begin{align}
        [X_t^L, h_t, c_t] = LSTM(H_t^C, h_{t-1}, c_{t-1}),
    \end{align}
    \begin{align}
        H^L_t = MLP_L(X_t^L) \in \mathbb{R}^{64}.
    \end{align}

     We use the LSTM block to remember which areas each newly deployed radar covers during the deployment process. The vector $H^L_t$ is then used to guide the actor network in radar placement decisions and the critic network in obtaining rewards.

     \end{itemize}
    
\subsubsection{The Actor and Critic Network}

After the LSTM network outputs the embedding vector $H^L_t$, we concatenate it with the remaining information from the MDP state vector, which contains $J, \, H_t ,\, v_t$, and can be detailed as follows:
\begin{align}
X_t^{P} = Concat(H^{L}_t, \, J, \, R_t ,\, v_t) \in \mathbb{R}^{64 + 2\times |J| + 2\times |R| + |R|}.
\end{align}

The concatenated vector $X_t^{P}$ serves as the input to the actor and critic networks. We use an MLP for both actor and critic network to obtain the action and the value function. The hidden layer size for both the actor and critic networks is 64, and the activation function is the sigmoid function. We use the following formulas to describe the outputs of the actor and critic networks:
\begin{align}
[\mu_t, \sigma_t] = MLP_{a}(X_t^{P}),
\end{align}
\begin{align}
V^\pi(X_t^P) = MLP_{v}(X_t^{P}).
\end{align}

Finally, the parameterized policies can be detailed as follows:
\begin{align}
\pi_\theta(a_t|X_t^P) = \frac{1}{\sqrt{2\pi} \sigma_t} e^{-\frac{(a_t-\mu_t)^2}{2\sigma_t^2}}.
\end{align}

We use PPO algorithm \cite{schulman2017proximal} to train and Adam optimizer \cite{kingma2014adam} to optimize our model. The detailed information of PPO algorithm can be found in Appendix \ref{sec:PPO}.

\subsection{Reward Based on Constraint Violation Degree Penalty and Exponential Function}
\label{sec:rewardshaping}

At time step $t$, when the agent receives a reward $r_t \in \mathcal{R}$, we perform two operations (namely CVDP-EXPR) to enable more efficient learning (since $B_R'$ is a broken line, which will be inconvenient for further discussion, we straighten $B_R'$ into a segment):

\begin{itemize}

    \item Constraint Violation Degree Penalty (CVDP): An intuitive idea is that uniformly deploying radars means each radar can cover the target area more evenly. Therefore, we designed CVDP based on the constraint violation. First, suppose radars are uniformly placed on $B_R$, with their position set denoted as $L = \{l_1, \cdots, l_{|R|}\}$. Next, we define a threshold set $U = \{u_1, \cdots, u_{|R|}\}$ and guide the agent to deploy the radar within the set $B(l_t, u_t) \cap B'_R$ at time step $t$, where $B(x,r)$ denotes a line segment centered at $x$ with length $2r$. Formally, assume the agent takes action $a_t$, we designed the following constraint violation degree penalty:
    \begin{equation}
    \begin{split}
        p_t = \begin{cases}
        0 & \text{if} \; |l_t - a_t| \le u_t, \\
        u_t - 3 \times |l_t - a_t| & \text{if} \; u_t < |l_t - a_t| \le 3/2u_t, \\
        u_t - |l_t - a_t| & \text{if} \; |l_t - a_t| > 3/2u_t.
        \end{cases}
    \end{split}
    \end{equation}

    \item Exponential Function Reward (EXPR): We found that after deploying the first few radars, they could cover the vast majority of jamming-free areas. This resulted in diminishing returns when placing new radars, as the additional coverage gain became smaller. Consequently, reward fluctuations became less pronounced, leading to inefficient learning by the agent in subsequent time steps.

    To address this, we designed the EXPR: Given that the environment provides a reward $r_t$ at time step $t$ and a reward $r_{t-1}$ at the previous time step, we formulate the exponential-like reward as follows:
    \begin{align}
        r_t' = \frac{10^{r_{t}} - 10^{r_{t-1}}}{10}.
    \end{align}
\end{itemize}

Finally, the reward $\tilde{r}_t$ guiding the agent's update is defined by the following equation:
\begin{align}
    \tilde{r}_t = r_t' + p_t.
\end{align}

We can chain all the networks together and obtain a pseudocode for network training in FARDA, represented by Algorithm \ref{algorithm}.

\begin{algorithm}
\caption{The Pseudocode for Network Training in FARDA}
\label{algorithm}
\KwIn{Encoder network ${\psi}$, actor network $\theta$, critic network $\phi$, learning rates $\eta_\psi, \, \eta_\theta, \eta_\phi$, total episodes $E$, training steps $T_{train}$, required deploy count of radars $T$, replay buffer $B$}
\For {$e = 1$ to $E$ \do }{
    Generate initial state $s_0 = \{M_0,\, J,\, H_0, v_0 \}$ from environment\;
    Using eq (\ref{eq:lstminit}) to init LSTM begin state $h_0, \, c_0 $\;
    $B \leftarrow \{\}$ \;
    \For {$t = 0$ to $T - 1$ \do }{
        Use $M_t$ to calculate $M'_t$ using equation \eqref{eq:one more channel}\;
        $H^L_t\leftarrow $\textbf{Enc}$([M_t, M'_t])$\;
        $X_t^P = $ Concat$(H_t^L,\, J,\, R_t,\, v_t) $ \;
        $[\mu_t, \sigma_t] \leftarrow $ \textbf{Actor}$(X_t^{P})$\;
        Sample $a_t$ from $\mathcal{N}(\mu_t,\, \sigma_t^2)$\;
        $v^\pi(X_t^{P}) \leftarrow $ \textbf{Critic}$(X_t^{P})$\;
        Get reward $r_t$, next state $s_{t+1}$ from environment using action $a_t$\;
        Perform CVDP-EXPR to get $\tilde{r}_t$ \;
        $B \leftarrow$ \textbf{Insert}$(s_t,\, a_t,\, \tilde{r}_t,\, s_{t+1})$ \;
    }
    \For {$t = 0$ to $T_{train}$ \do }{
        Perform PPO algorithm using buffer $B$ with batch $T$ using Adam optimizer with learning rate $\eta_\psi,  \, \eta_\theta, \eta_\phi$\;
    }
}
\end{algorithm}

\section{Experiments}

To validate the effectiveness of FARDA, we design numerical experiments tailored to specific environments. We compare FARDA with the PSO1D and GA1D algorithms on the test dataset. The results demonstrate that FARDA achieves coverage comparable to PSO1D and GA1D while operating nearly 7,000 times faster. Additionally, we conduct a series of ablation tests, whose outcomes confirm the essential role of each component within FARDA.

\subsection{Experimental Setups}

\label{sec:experiments setups}
We display the necessary setups of the experiment, detailed as follows:

\begin{itemize}
    \item Test Dataset: Due to the high computational cost of evolutionary algorithms, we randomly select 500 jamming node combinations from our environment as our test set. During training, we also select jamming node combinations randomly from the environment. This ensured the test and training datasets were sampled from the same distribution.

    \item Baselines: We use PSO1D algorithm and GA1D as our baselines. These two algorithms are based on PSO and GA, which are thoroughly introduced in Appendix \ref{sec:GA} and \ref{sec:PSO}.

    \item Evaluation metrics: We primarily use average coverage and computation time for evaluation. Specifically, because there is no efficiency metric to evaluate radar deployment, we designed a new metric for benchmarking deployment efficiency.

    \item Hardware configuration: Our policy network in FARDA is trained and tested on a Nvidia GeForce RTX 3090 GPU. We utilize a 12th Gen Intel® Core™ i9-12900 CPU for evolutionary algorithms to get the result.

    \item The number of jamming nodes and radars: In the environment, the number of enemy jamming nodes is 3, and we need to deploy four radars, i.e., $|J| = 3$,  $|R| = 4$.

    \item Hyperparameters: Our experiment contains three parts of hyperparameters: environment, evolutionary algorithms, and policy network in FARDA. 
    
    \begin{enumerate}
        \item For the environment after simplification: We set $\tilde{P}_r = 450\text{W}, \, \tilde{P}_t = 30\text{W},\, \lambda = 0.3\text{m}, \, K = 32, \, B = 10^6\text{Hz}, \, F_r = 2\times 10^3, \, T_0 = 270\text{K}, \, Pr_{fa} = 10^{-3}$, \, $F_e = 10^{3/10}.$ Besides, we set the number of pulses accumulated during the phase-correlation period $N = 16$, and we set $ \theta = 2\times 0.886 / N, G_T = {2\pi d_E (N-1)}/{\lambda}$, and $ G_R = 2\pi d_E/\lambda$. The Boltzmann constant is $K_e=1.38\times 10^{-23}$. Finally, we set the detection probability threshold $\tau$ to $0.5$. 

        \item For evolutionary algorithms: For GA1D, we set $Pr_c = 0.9,\, Pr_m = 0.1,\, T_{G} = 100 $, and $N_{G} = 50$. For PSO1D, we set $T_{P} = 100$, $N_{P} = 20$, $\omega = 1$, $c_1 = c_2 = 2.$

        \item For policy network in FARDA: The training rates are set to $\eta_\psi = 10^{-4},  \, \eta_\theta= 10^{-4}, \eta_\phi =  5\times10^{-4}$, the number of episodes $E = 2\times 10^5$,  the number of training steps in each iteration $T_{train} = 10$, and the clip ratio $\epsilon$ in PPO is set to $0.2$.
    \end{enumerate}
\end{itemize}

\subsection{Main Results}
For the dataset we have obtained, we compared FARDA with GA1D and PSO1D algorithms.

To better compare the efficiency of these algorithms, we specially designed an efficiency metric to measure the detected coverage in every log unit of time:
\begin{align}
    \label{eq:eff}
    efficiency = \frac{coverage}{\log(1+time)},
\end{align}
where we consider the impact of the scale of time becomes less significant as the deployment takes longer.

Table \ref{tab:main} shows the main testing results. It can be demonstrated that, compared to the evolutionary algorithms GA1D and PSO1D, our FARDA exhibits superior performance in all metrics. The most significant improvement is in the time metric, where FARDA is approximately 7000 times faster than evolutionary algorithms, reducing the computation time from about half an hour to less than half a second. Also, FARDA significantly enhances the efficiency metric from 0.13 to 4.21. As for the least improved metric coverage, FARDA still outperforms evolutionary algorithms.

\begin{table}
\centering
\caption{Main result of FARDA and evolutionary algorithms: The time metric is the average response time. The \underline{underlined} parts indicate the best results.}
\begin{tabular}{c|ccc}
    \hline   Methods & Coverage(\%) & Time & Efficiency \\   
    \hline   PSO1D & 93.75 & 24.9m & 0.13 \\ 
             GA1D & 93.79 & 30.7m & 0.12  \\  
    \hline   FARDA & \underline{93.94} & \underline{0.25s} & \underline{4.21}  \\      
    \hline  
\end{tabular}
\label{tab:main} 
\end{table}

Additionally, we divide the instances in the test dataset into three categories: Bad, Normal, and Good. The dividing standard is the coverage of evolutionary algorithms: we define bad data as coverage lower than 0.9 when using both GA1D and PSO1D. For the normal data, the coverage threshold is set to be greater than 0.9 but lower than 0.95, while the rest of the data in the test dataset is considered as good data. We test FARDA on these three categories and compare with evolutionary algorithms. The results are shown in Table \ref{tab:sub}.

\begin{table}
\centering
\caption{Average coverage (\%) comparison on different datasets. The improvement ratio is FARDA compared with GA1D, and the \underline{underlined} parts indicate the best results.}
\begin{tabular}{c|ccc}
    \hline    Data Type & Bad Data & Normal Data & Good Data  \\   
    \hline     Count & 56 & 288 & 156 \\
    \hline   PSO1D & 89.31 & 92.90 & 96.92  \\ 
             GA1D & 89.44 & 92.94 & 96.92 \\  
    \hline   FARDA & \underline{89.70} & \underline{93.10} & \underline{97.00} \\   
 
    Improvement Ratio & 0.29\% & 0.17\% & 0.08\%\\
    \hline   
\end{tabular}

\label{tab:sub}
\end{table}

We found that in these three categories, FARDA outperforms evolutionary algorithms in all three datasets. Furthermore, higher improvements are shown in bad data and normal data. In real-world military scenarios, the enemy tends to make the jamming node combination more difficult to detect targets, making our method more useful in real-life applications.

\subsection{Ablation Tests}

To demonstrate the effectiveness of each part of FARDA, we designed two parts of ablation tests focused on network design and reward design, respectively. Since the forward propagation time of the following networks is approximately the same, as reported in \ref{tab:main}, we do not record time and efficiency metrics.

First, we design ablation tests targeting the network by removing the CNN, LSTM, and the whole encoder, respectively. The results are shown in Table \ref{tab:ablation1}:

\begin{table}
\centering
\caption{The ablation test of our network design in FARDA. We report the average coverage of each part in the test dataset. The \underline{underlined} part indicates the best result.}
\begin{tabular}{c|cccc}
    \hline   Methods &  
FARDA & w/o CNN &  w/o LSTM & w/o Encoder \\   
    \hline   Coverage(\%) 
     & \underline{93.94} &  93.89 & 93.90 & 91.44  \\ 
    \hline   
\end{tabular}

\label{tab:ablation1}
\end{table}

It can be observed that our proposed encoder in FARDA performs best. In addition, removing either the CNN or the LSTM leads to a decrease in performance, and removing the entire Encoder results in a significant drop in average coverage. Upon analysis, we found that without the encoder, all the high-dimensional heatmap information is fed directly to the actor network, while the hidden size of the actor is relatively small compared with the dimension of the heatmap. This may prevent the actor from extracting the necessary and correct information from the heatmap, leading to incorrect decision-making.

Next, we explored the impact of CVDP-EXPR. We conducted ablation experiments by removing CVDP, EXPR, and CVDP-EXPR. The results are shown in Table \ref{tab:ablation2}:

\begin{table}
\centering
\caption{The ablation test of our reward design. We report the average coverage of each part in the test dataset. The \underline{underlined} part indicates the best result.\label{tab:ablation2}}
\begin{tabular}{c|cccc}
    \hline   Methods & FARDA &  w/o CVDP & w/o EXPR &  w/o CVDP-EXPR \\   
    \hline   Coverage(\%) & \underline{93.94} & 88.80 & 93.90 & 88.80\\ 
    \hline   
\end{tabular}
\end{table}

\begin{figure}[!t]
\centering
\subfloat[\centering FARDA]{\includegraphics[width=0.45\textwidth]{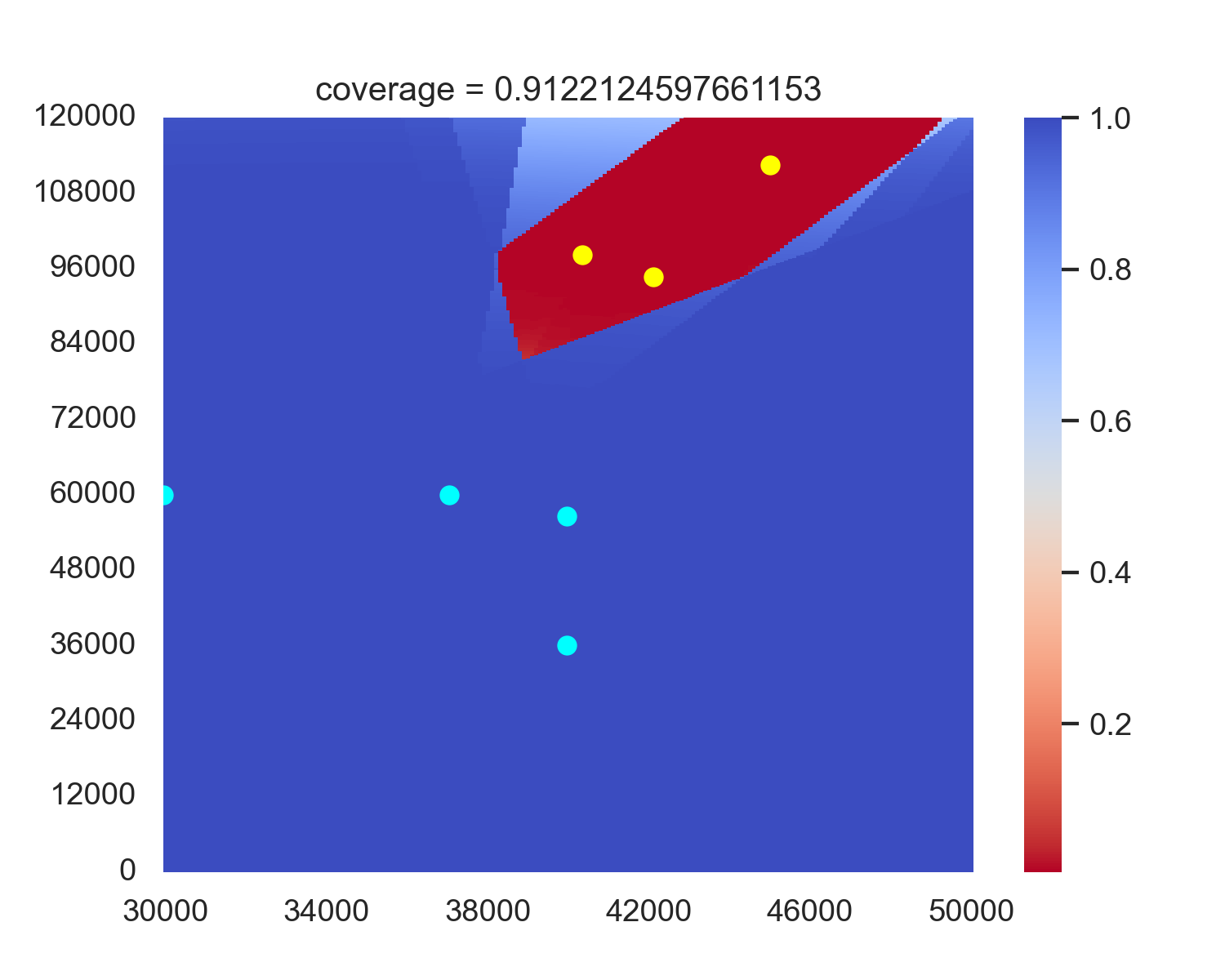}}
\subfloat[\centering w/o CVDP]{\includegraphics[width=0.45\textwidth]{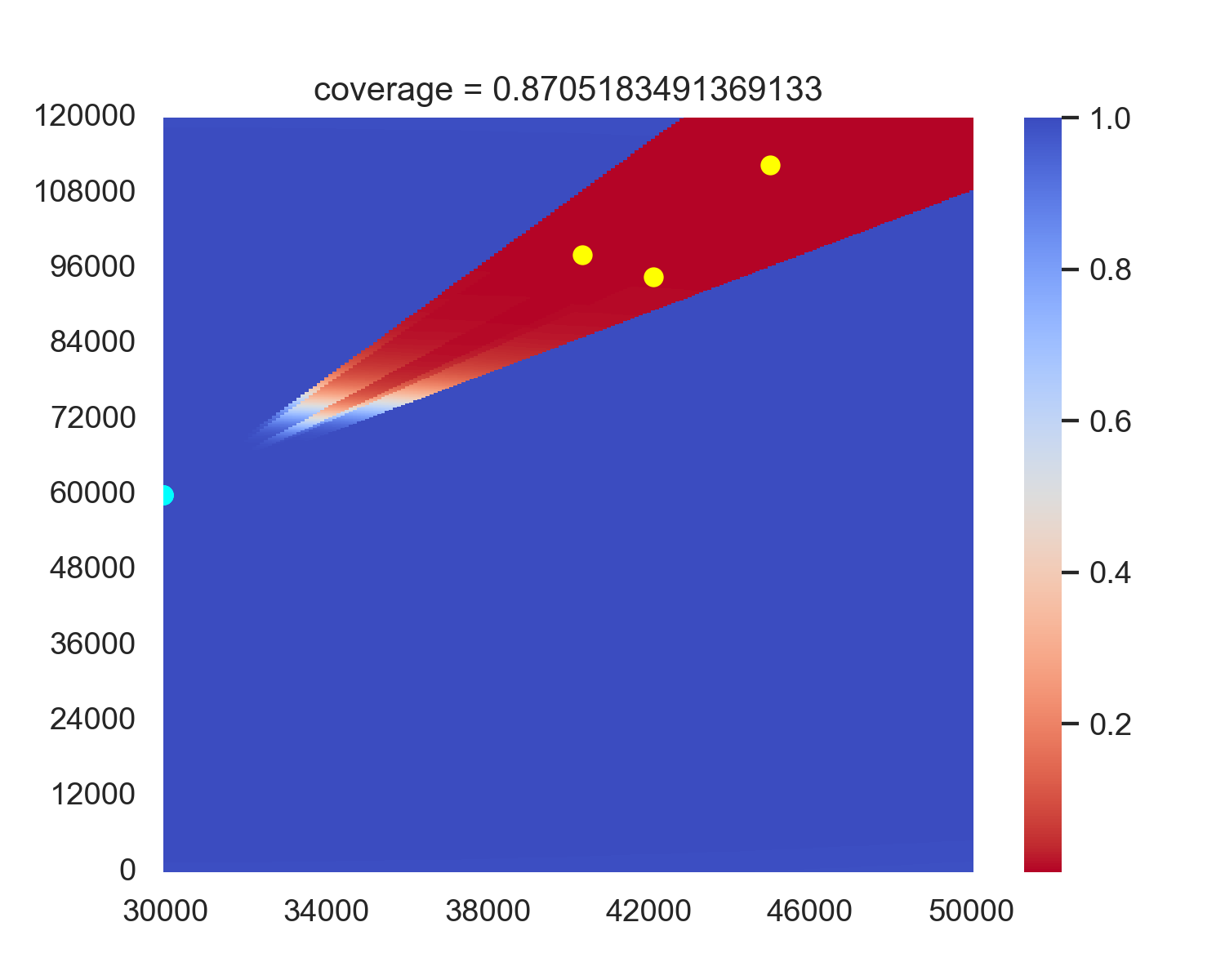}}
\caption{(a) An example radar deployment for FARDA, (b) An example of removing CVDP in FARDA. The result of removing CVDP-EXPR is the same as (b). \label{fig:ablation2}}
\end{figure}

It can be seen that FARDA shows the best performance, and deleting any other part leads to varying degrees of decline in coverage. Notably, the removal of the CVDP results in a significant performance drop. Our analysis revealed that without CVDP, the training collapses: As shown in Figure \ref{fig:ablation2}, all radars tend to move in the same direction and get stuck. This phenomenon also occurs when removing CVDP-EXPR, as it involves the removal of CVDP. Further analysis indicates that removing CVDP causes the agent to consistently receive positive rewards, which gradually increases the probability of selecting incorrect actions, and the agent is unable to learn the correct way to place radars. However, with CVDP, if the agent deviates too far from the specified position, it receives punishment, which means the reward can be negative, reducing the probability of the agent selecting those actions and thus preventing training collapse.

\section{Conclusion}
We introduced a brand new approach, FARDA, in the anti-jamming situation of cognitive radar deployment: a novel pipeline combining the encoder network and reinforcement learning. We have shown that FARDA has significantly improved deployment time, while coverage also outperforms evolutionary algorithms. The improvement in time shows the prospect of utilizing our approach in real-world scenarios, where fast deployment is as important as coverage. 

Our subsequent research will focus on two aspects: more complex static environments and dynamic environments. In the more complex static environment section, the number of radars and jamming nodes will be adjusted accordingly, while the size of region $S$ and the shape of the deployable area will also change. In the dynamic environment section, the positions of jamming nodes may shift over time, requiring a dynamic response of radars to their positions. However, unlike the static scenario, each radar position at any given moment cannot deviate too far from its previous position. Additionally, the evaluation metrics will also change.

\section*{Acknowledgment}

This paper was supported by National Key R\&D (research and development) Program of China (2021YFA1000403). We want to thank Wenzhao Liu, Chi Zhou, and Wang Luo for their valuable discussion. We would also like to thank reviewers for their valuable comments.

\bibliographystyle{unsrt}  
\bibliography{references}  






\newpage

\section*{Appendix}

\appendix

\section{Genetic Algorithm}

\label{sec:GA}
Genetic Algorithm: The Genetic Algorithm (GA) was introduced by \cite{holland1992genetic} first and was initially designed as a computational model to simulate the natural process of evolution. It uses computer simulations of chromosome crossover and mutation processes to find the optimal solution.

Formally, suppose we have the following problem to optimize:
\begin{align}
\label{eq:appendix}
\begin{split}
    \max& \quad f(x) \\
    s.t.& \quad x\in D\subset \mathbb{R}^{n}
\end{split}
\end{align}

Then, GA initialize a set of chromosomes $X^{0} = \{x_1^{0}, x_2^{0}, \cdots, x_{N_G}^{0}\}, \, x_i \in D$. And calculate their fitness value $A^0 = \{f(x_1^0),f(x_2^{0}), \cdots, f(x_{N}^{0}) \}$. Then, during every iteration, we perform certain steps; in each step, we conduct selection, crossover, and mutation operations in order.

In each step of iteration $t$, we randomly select two chromosomes $x_i^{t-1}$ and $x_j^{t-1}$ with the probability of their fitness value $\{f(x_i^{t-1}) / \sum_j{f(x_j^{t-1})}\}$. Then, these two chromosomes have a probability of $Pr_c$ to perform the following crossover operation:
\begin{align}
\label{eq:crossover}
\begin{split}
    \hat{x}_i^{t} = k\times x_i^{t-1} + (1-k)x_j^{t-1} ,\\
    \hat{x}_j^{t} = k\times x_j^{t-1} + (1-k)x_i^{t-1},
\end{split}
\end{align}

where $k$ is randomly selected in $[0,1]$. Besides, if $\hat{x}_i^{t} \notin D$, we see the crossover operation fails and do not perform it.

After crossover operation, we got two vectors: $\hat{x}_i^{t}$ and $\hat{x}_j^{t}$. Then we will perform the mutation operation: for each chromosome, it has $Pr_m$ probability to become a new random vector in $D$.

Finally, after these two operations, we can obtain the new vectors $x_i^{t}$ and $x_j^{t}$, and we append these vectors to the new population $X^t$. We then repeatedly perform these steps until the size of $X^t$ reaches $N_G$. We finally calculate the fitness of the new population and start a new iteration.

\section{Particle Swarm Optimization Algorithm}

\label{sec:PSO}

The Particle Swarm Optimization Algorithm was proposed by \cite{kennedy1995particle}, originally designed to simulate the movement of flocks of birds. Its core principle involves dynamically adjusting the search direction through individual memory (historical optimum) and collective information sharing (global optimum) to locate the optimal solution.

Formally, suppose the optimization problem is still \eqref{eq:appendix}. In the PSO algorithm, we need to set the iteration number $T_P$ and the population size $N_P$. We will initialize the position and speed of the particle population:
\begin{align}
\begin{split}
    X^0 = (x^0_1,\,x^0_2,\, \cdots, \,x^0_{N_P}) &,\\
    V^0 = (v^0_1,\,v^0_2,\, \cdots, \,v^0_{N_P}) &.
\end{split}
\end{align}

Then, we will keep a record of the best position of every particle $P = (p_{1}, \cdots, p_{N_P})$, along with the global best position $g$. 

Then, in iteration $t$, we will update the speed and position of every particle using the following formula:
\begin{align}
\begin{split}
    v_{ti}& = \omega * v_{t-1, i} + c_1 r_1(p_i - x_{t-1, i}) + c_2r_2 (g - x_{t-1, i}), \\
    x_{ti} &= \begin{cases}
    x_{ti} + v_{ti} & \text{if } (x_{ti} + v_{ti})\in D ,\\
    x_{ti} & \text{else},
\end{cases}
\end{split}
\end{align}

where $\omega$ is inertia weight, $c_1$ and $c_2$ are acceleration constants, $r_1$ and $r_2$ are selected randomly in interval $[0,1]$.

After the update of every particle, we can then update the best positions $P$ and $g$:
\begin{equation}
\begin{split}
    p_i &= \max(p_i, x_{ti}) , \, i = 1,2,\cdots N_P, \\
    g &= \max(g, \max_{i=1} ^ {N_p} x_{ti}).
\end{split}
\end{equation}

\section{Proximal Policy Optimization Algorithm}

\label{sec:PPO}

We use PPO algorithm to train our model. PPO algorithm was first introduced by \cite{schulman2017proximal}, and was optimized from the Trust Region Policy Optimization \cite{schulman2015trust} algorithm by using only first-order optimization. The author shows that PPO algorithm performs better in continuous action spaces. PPO algorithm optimizes agents by maximizing the following surrogate objective:
\begin{equation}
\label{eq:ppo}
\begin{split}
    \mathbb{E}_t[\min(\frac{\pi_\theta(a_t|s_t)}{\pi_{\theta_{old}}(a_t|s_t)}A^t,\,\text{clip}(\frac{\pi_\theta(a_t|s_t)}{\pi_{\theta_{old}}(a_t|s_t)},\, 1-\epsilon, \, 1 +\epsilon)A^t].
\end{split}
\end{equation}

In \eqref{eq:ppo}, $\pi_\theta$ and $\pi_{\theta_{old}}$ are the current policy model and the policy model before update. $A^t$ is the advantage function, which is computed using Generalized Advantage Estimation \cite{schulman2015high}.

\end{document}